\documentclass{article} 
\usepackage[preprint]{colm2026_conference}

\usepackage{microtype}
\usepackage{hyperref}
\usepackage{url}
\usepackage{booktabs}
\usepackage{amsmath}
\usepackage{amssymb}
\usepackage{graphicx}
\usepackage{multirow}
\usepackage{subcaption}
\usepackage{float}

\usepackage{lineno}

\definecolor{darkblue}{rgb}{0, 0, 0.5}
\hypersetup{colorlinks=true, citecolor=darkblue, linkcolor=darkblue, urlcolor=darkblue}

\title{Does Your Optimizer Care How You Normalize? \\ Normalization-Optimizer Coupling in LLM Training}

\author{Abdelrahman Abouzeid \\
Georgia Institute of Technology \\
\texttt{aabouzeid3@gatech.edu}}

\begin{document}

\ifcolmsubmission
\linenumbers
\fi

\maketitle

\begin{abstract}
In LLM training, normalization layers and optimizers are typically treated as independent design choices. In a $3 \times 2$ factorial at 1B parameters and 1000 training steps, we show this assumption can fail: Dynamic Erf \citep[Derf;][]{chen2026strongernormalizationfreetransformers} suffers a large negative interaction with Muon \citep{jordan2024muon}, with its gap to RMSNorm growing from $+$0.31 nats under AdamW to $+$0.97 under Muon, approximately three times larger. Dynamic Tanh \citep[DyT;][]{Zhu_2025_CVPR}, included as a bounded-normalizer control, shows no such penalty. Our evidence points to two failure modes of erf under Muon's faster spectral-norm growth: saturation (lossy compression) and scale blindness (discarding activation magnitude). An EMA-blend that reintroduces running scale estimates recovers ${\sim}84\%$ of the gap. Separately, reducing Derf's $\alpha$ from its published default (0.5 to 0.3) recovers ${\sim}80\%$ by keeping erf in its near-linear regime, where it approximately preserves relative scale; this setting is not the published default of \citet{chen2026strongernormalizationfreetransformers}. Using Derf's published default $\alpha$ with Muon incurs a 0.66-nat interaction penalty without producing NaNs or divergence, making the failure easy to miss in short pilot runs.
\end{abstract}

\section{Introduction}
\label{sec:intro}

\emph{Normalization-free architectures} replace the LayerNorm/RMSNorm layers standard in transformers \citep{NIPS2017_3f5ee243} with pointwise bounding functions such as Dynamic Erf \citep[Derf;][]{chen2026strongernormalizationfreetransformers}, which uses $\gamma \cdot \mathrm{erf}(\alpha x + s) + \beta$. By eliminating the reduction operation (computing RMS across the hidden dimension), these methods promise faster kernels and, crucially for distributed training, elimination of the cross-device allreduce required by RMSNorm under tensor parallelism. Independently, \emph{Muon} \citep{jordan2024muon}, a spectral optimizer, replaces AdamW's \citep{loshchilov2018decoupled} per-parameter adaptive scaling with Newton-Schulz orthogonalization, showing strong results on standard RMSNorm architectures. In current LLM practice, these two choices are often treated as modular.

That modularity assumption has not been tested where it matters most. The Derf and DyT papers use AdamW exclusively, while the Muon and Moonlight papers use RMSNorm exclusively. To our knowledge, no prior work evaluates these components jointly. We therefore study a controlled $3 \times 2$ factorial, $\{\text{RMSNorm}, \text{Derf}, \text{DyT}\} \times \{\text{AdamW}, \text{Muon}\}$, at 1B parameters, with DyT included as a second bounded normalizer to test whether any incompatibility is Derf-specific or generic to the class.

Our central finding is that \textbf{these design choices do not behave independently in the tested regime}. Derf's gap to RMSNorm is \emph{approximately three times larger} under Muon than under AdamW ($+$0.97 vs.\ $+$0.31 nats\footnote{Throughout, ``nats'' refers to cross-entropy loss with natural logarithm.} at 1000 steps), whereas DyT shows no such negative interaction ($-$0.10). The coupling is Derf-specific.

\begin{figure}[t]
\centering
\includegraphics[width=0.82\columnwidth]{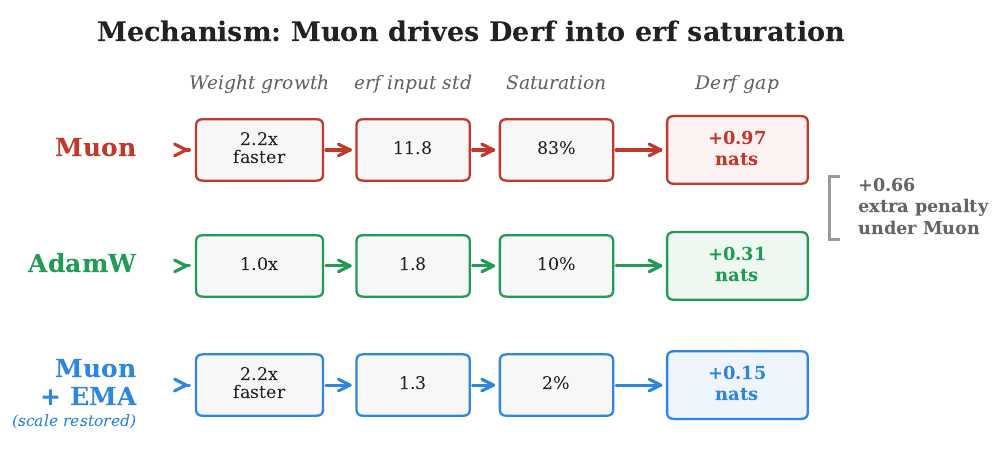}
\caption{Overview of the optimizer-driven saturation mechanism. At the tested learning rates, Muon grows weights ${\sim}2.2\times$ faster than AdamW, pushing Derf into steep saturation (83\% of elements at $\pm 1$). AdamW's slower growth keeps Derf closer to its useful regime (10\% saturation). Muon+EMA (bottom) preserves Muon's fast weight growth but uses a running $\hat{\sigma}$ to keep the post-blend erf argument near unit scale (std 1.3, 2\% post-blend saturation), recovering ${\sim}84\%$ of the Derf penalty. Weight and saturation numbers from layer 15, step 950 (seed 42); quality gaps at step 1000.}
\label{fig:mechanism}
\end{figure}

The failure mode is subtle but consequential. Derf+Muon trains to completion with no NaN or gradient explosion, yet its loss gap to RMSNorm+Muon widens progressively from step ${\sim}200$ onward. Our diagnostics point to erf's steep saturation profile: erf reaches 99\% output at $|\alpha x|\!=\!1.82$ (vs.\ $2.65$ for tanh), i.e.\ $|x| > 3.6$ at $\alpha\!=\!0.5$, and at the tested learning rates Muon grows weights ${\sim}2\times$ faster than AdamW, pushing Derf into a regime where distinct inputs collapse to identical $\pm 1$ outputs (Figure~\ref{fig:mechanism}).

We make the following contributions:

\begin{enumerate}
    \item \textbf{Discovery: a progressive, no-crash Derf-Muon incompatibility.} In a $3 \times 2$ factorial at 1B (3 seeds/cell), Derf's penalty under Muon ($+$0.97) is ${\sim}3\times$ larger than under AdamW ($+$0.31), while DyT shows no such coupling ($-$0.10 interaction). The phenomenon is specific to Derf, not generic to bounded normalizers.

    \item \textbf{Mechanism: the evidence points to saturation and scale blindness as joint failure modes.} At the tested learning rates, Muon produces ${\sim}8\times$ more erf saturation than AdamW (83\% vs.\ 10\% at layer 15, step 950). Reducing $\alpha$ from 0.5 to 0.3 (non-default) recovers ${\sim}80\%$ of the gap by keeping erf in its near-linear regime, where it approximately preserves relative scale. EMA-blend recovers a comparable ${\sim}84\%$ via running scale estimates (Section~\ref{sec:ablations}). asinh pre-compression, which eliminates saturation without restoring scale, recovers ${\sim}49\%$.

    \item \textbf{Practical implication: normalization-free methods benefit from optimizer-aware validation.} The incompatibility can be addressed by $\alpha$ tuning (zero overhead) or EMA-blend (0.15 nats residual gap, $128\times$ fewer allreduce operations than RMSNorm, $7.8\times$ norm-layer speedup at 8-way TP). However, fixed $\alpha$ may not scale to longer training horizons where residual magnitudes continue to grow (Section~\ref{sec:limitations}).
\end{enumerate}

\section{Background}
\label{sec:background}

\textbf{RMSNorm} \citep{NEURIPS2019_1e8a1942} normalizes by the root mean square: $\mathrm{RMSNorm}(x) = \gamma \cdot x / \sqrt{\mathrm{mean}(x^2) + \epsilon}$. This requires a reduction but is \emph{lossless}: if $x_i = 2 x_j$ before, then $x_i = 2 x_j$ after. Output has $\|x\|_{\mathrm{RMS}} \approx 1$.

\textbf{Derf} \citep{chen2026strongernormalizationfreetransformers} uses $\mathrm{Derf}(x) = \gamma \cdot \mathrm{erf}(\alpha x + s) + \beta$. No reduction needed, but erf is \emph{lossy}: when $|\alpha x| > 3$, inputs of magnitude 10 and 100 both map to $\pm 1$.

\textbf{DyT} \citep{Zhu_2025_CVPR} uses $\mathrm{DyT}(x) = \gamma \cdot \tanh(\alpha x) + \beta$ with separate $\alpha$ per sublayer. Like Derf, it is reduction-free and bounded, but lacks the learnable shift $s$.

\textbf{Muon} \citep{jordan2024muon} orthogonalizes the gradient for 2D weights: $U = \mathrm{NS}(G)$, $W \leftarrow W - \eta U$. Updates have fixed spectral norm; unlike AdamW, where $\sqrt{v_t}$ dampens large gradients.

\textbf{EMA-blend} (our causal probe) keeps one running activation-scale estimate $\hat{\sigma}$ per normalization site, updates it once per training step, and blends raw and normalized inputs:
\begin{equation}
    \mathrm{Derf\text{-}EMA}(x) = \gamma \cdot \mathrm{erf}\!\left(\alpha \left[(1\!-\!\lambda)\, x + \lambda\, \frac{x}{\hat{\sigma}}\right] + s\right) + \beta
    \label{eq:ema_blend}
\end{equation}
where $\hat{\sigma}$ is a running EMA of activation std. The normalized term restores scale information, while the raw term preserves a small growth brake. We use mixing coefficient $\lambda = 0.9$ and smoothing factor $m = 0.5$ (Section~\ref{sec:ema}).

\section{Experimental Setup}
\label{sec:setup}

We use the Llama architecture \citep[LlamaForCausalLM;][]{touvron2023llama2openfoundation} (GQA, SwiGLU, RoPE) at 1B parameters (2048 hidden, 16 layers, 32 heads, 8 KV heads, 5632 intermediate, 33 norm instances). Training: FineWeb-Edu \citep{penedo2024finewebdatasetsdecantingweb}, Llama 3.2 tokenizer \citep{DBLP:journals/corr/abs-2407-21783} (128K vocab), sequence length 2048, micro-batch 8, gradient accumulation 64 ($\sim$1M tokens/step), 1000 steps ($\sim$1B tokens), cosine decay, 100-step warmup. AdamW uses LR $3 \times 10^{-4}$, $(\beta_1, \beta_2) = (0.9, 0.95)$, and weight decay $0.1$ (applied to 2D parameters only); Muon uses LR 0.02 with no weight decay. Gradient clipping max\_norm$=$1.0. Training uses bfloat16 autocast/model weights; all scalar normalization parameters ($\alpha$, shift $s$, $\gamma$, $\beta$) and the Derf/DyT normalization forward pass remain in float32 for stability; this applies to \emph{all} runs, including every intervention in Table~\ref{tab:fixes}. Hardware: single NVIDIA H200 SXM.

We test $\{\text{RMSNorm}, \text{Derf}, \text{DyT}\} \times \{\text{AdamW}, \text{Muon}\}$ plus EMA-blend+Muon. Derf uses $\alpha=0.5$, $s=0$ \citep{chen2026strongernormalizationfreetransformers}; DyT uses $\alpha_{\mathrm{attn}}=0.5$, $\alpha_{\mathrm{ffn}}=0.3$ \citep{Zhu_2025_CVPR}. All configs run with seeds 42, 43, 44. All main-paper runs train to completion (${\sim}$250 GPU-hours; training on H200 SXM, TP benchmark on H100 NVLink). Every 10 steps, we log per-layer diagnostics: activation RMS, erf saturation fraction, erf input std, weight norms, and alpha gradients. The main factorial, mechanism, and EMA results use the 1000-step runs. Table~\ref{tab:fixes} summarizes targeted ablation probes, with per-intervention details in Appendix~\ref{app:fixes}; Exploratory 125M pilots appear in Appendix~\ref{app:125m}.

\section{Results}
\label{sec:results}

\subsection{The $3 \times 2$ Factorial}

\begin{table}[t]
\centering\small
\begin{minipage}[t]{0.44\linewidth}\centering
\begin{tabular}{lcc}
\toprule
 & \textbf{AdamW} & \textbf{Muon} \\
\midrule
\textbf{RMSNorm} & 3.883\,{\scriptsize$\pm$\,0.008} & \textbf{3.322}\,{\scriptsize$\pm$\,0.003} \\
\textbf{Derf} & 4.192\,{\scriptsize$\pm$\,0.003} & 4.289\,{\scriptsize$\pm$\,0.017} \\
\textbf{DyT} & 4.201\,{\scriptsize$\pm$\,0.008} & 3.541\,{\scriptsize$\pm$\,0.003} \\
\bottomrule
\end{tabular}
\end{minipage}%
\hfill
\begin{minipage}[t]{0.54\linewidth}\centering
\begin{tabular}{lccc}
\toprule
 & \textbf{AdamW gap} & \textbf{Muon gap} & \textbf{Interaction} \\
\midrule
\textbf{Derf} & $+$0.31 & $+$0.97 & $\mathbf{+0.66}$ \\
\textbf{DyT} & $+$0.32 & $+$0.22 & $\mathbf{-0.10}$ \\
\bottomrule
\end{tabular}
\end{minipage}
\caption{\emph{Left:} Validation loss at 1000 steps (1B params, 3 seeds/cell). \emph{Right:} Gap to RMSNorm under each optimizer; interaction $=$ Muon gap $-$ AdamW gap. Under AdamW, Derf and DyT have near-identical gaps ($+$0.31, $+$0.32). Under Muon, Derf's gap triples ($+$0.66 interaction) while DyT's shrinks ($-$0.10). Bootstrap 95\% CIs: Derf $[0.64, 0.68]$, DyT $[-0.12, -0.08]$ (Appendix~\ref{app:stats}).}
\label{tab:factorial}
\end{table}

Table~\ref{tab:factorial} reveals an asymmetry. Under AdamW, Derf and DyT have similar penalties ($+$0.31, $+$0.32). Under Muon, they diverge sharply: Derf's gap triples to $+$0.97, while DyT's gap \emph{shrinks} to $+$0.22. Derf+Muon ($4.289 \pm 0.017$) is worse than Derf+AdamW ($4.192 \pm 0.003$): Muon's 0.56-nat optimizer advantage is not merely negated but \emph{reversed} under Derf. In contrast, Muon benefits DyT even more than RMSNorm (0.66 vs.\ 0.56 nats). The interaction is Derf-specific ($+$0.66), not inherent to bounded normalizers (DyT: $-$0.10).

\subsection{The Coupling Is Derf-Specific and Progressive}
\label{sec:dyt_specific}

Despite comparable Muon-driven weight growth, DyT reaches only 27\% saturation vs.\ Derf's 79\% (Appendix Table~\ref{tab:dyt_diagnostics}). Two effects separate them: tanh's 99\% output occurs at $|x|\!=\!2.65$ vs.\ $1.82$ for erf (46\% wider useful regime), and DyT's $\alpha$ self-corrects more aggressively under Muon (42\% drop vs.\ 28\%) because tanh retains usable gradients at lower saturation. The coupling is therefore Derf-specific: Derf's narrow saturation geometry is incompatible with Muon's scale dynamics, while DyT avoids the problem.

The coupling also strengthens over training (Table~\ref{tab:interaction}).

\begin{table}[t]
\begin{center}\small
\begin{tabular}{lcc}
\toprule
\textbf{Step} & \textbf{Derf+Muon gap} & \textbf{EMA-blend gap ($\lambda\!=\!0.9$)} \\
\midrule
100 & $+$0.13 & $+$0.43 \\
300 & $+$0.71 & $+$0.35 \\
500 & $+$0.96 & $+$0.19 \\
700 & $+$1.01 & $+$0.16 \\
1000 & $+$\textbf{0.97} & $+$\textbf{0.15} \\
\midrule
\multicolumn{3}{l}{\emph{Interaction at 1000: Muon gap $+$0.967, AdamW gap $+$0.308 $\Rightarrow$ \textbf{+0.659}}} \\
\bottomrule
\end{tabular}
\end{center}
\caption{Gap to RMSNorm+Muon over training (3-seed averages). Derf+Muon widens and plateaus; EMA-blend ($\lambda\!=\!0.9$) narrows monotonically from $+$0.43 to $+$0.15. Both trends are partially confounded with cosine decay.}
\label{tab:interaction}
\end{table}

At step 100, the Derf gap is just $+$0.13; it widens $7\times$ over 900 steps, most steeply at near-peak LR (steps 100--300), consistent with weight-growth-driven saturation. The interaction is tight across seeds: mean $+0.659$, std $\leq 0.017$ (Appendix~\ref{app:stats}).

\section{Mechanistic Analysis}
\label{sec:mechanism}

\subsection{erf's Narrow Operating Regime}

\begin{figure}[t]
\centering
\includegraphics[width=0.95\columnwidth]{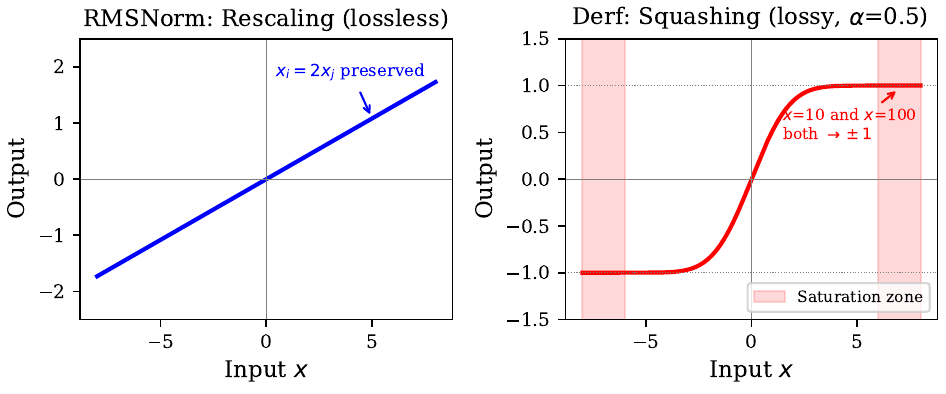}
\caption{Rescaling (RMSNorm) preserves relative magnitudes at any input scale; squashing (Derf) destroys them outside a narrow operating regime.}
\label{fig:squash_rescale}
\end{figure}

\textbf{Rescaling} (RMSNorm) is invertible and scale-invariant (Figure~\ref{fig:squash_rescale}): growing activations increase the divisor, maintaining unit output.

\textbf{Squashing} (Derf): erf maps $\mathbb{R} \to [-1, 1]$. Unlike RMSNorm's lossless rescaling, erf introduces two failure modes that the asinh and EMA-blend probes help disentangle:

\begin{itemize}
    \item \textbf{Saturation: lossy compression.} When $|\alpha x| > 3$, erf collapses to $\pm 1$, a many-to-one mapping where inputs of magnitude 10 and 100 become identical. This is a primary observed failure mode of \textbf{Derf+Muon}: erf input std reaches 11.8 at layer 15 by step 950 (Table~\ref{tab:weight_growth}), saturating 83\% of elements. No other configuration in our factorial reaches this level; Derf+AdamW stays at 10\% because, at the tested learning rates, Muon grows weights $2.2\times$ faster. asinh pre-compression (Appendix~\ref{app:fixes}) eliminates saturation entirely (0\% at all layers) and recovers ${\sim}49\%$ of the Derf-Muon gap (3.82 vs.\ 4.29 at step 900, single-seed follow-up).

    \item \textbf{Scale blindness: erf discards activation magnitude.} asinh eliminates saturation but not scale information, recovering only ${\sim}49\%$ vs.\ EMA-blend's ${\sim}84\%$; the 35\,pp gap is consistent with scale blindness as a distinct factor beyond desaturation alone. Lowering $\alpha$ to 0.3 (Section~\ref{sec:alpha03}) recovers ${\sim}80\%$ because in the near-linear regime ($\mathrm{erf}(\alpha x) \approx 2\alpha x/\sqrt{\pi}$) erf approximately preserves relative magnitude, addressing both failure modes with one parameter.
\end{itemize}

RMSNorm and EMA-blend avoid these failures by preserving scale information; lowering $\alpha$ delays these failures by widening the useful regime. Without one of these mechanisms, Muon's faster weight growth pushes activations out of erf's useful regime with no path to self-correction.

\subsection{Muon's Weight Growth Drives Saturation}

\begin{table}[t]
\begin{center}
\begin{tabular}{lccc}
\toprule
\textbf{Metric (L15, step 950)} & \textbf{Derf+Muon} & \textbf{Derf+AdamW} & \textbf{EMA-blend} \\
\midrule
MLP weight $\|W\|_F$ & 155.8 & 69.6 & 163.5 \\
erf input std & 11.8 & 1.8 & 1.3 \\
erf saturated (\%) & 83\% & 10\% & 2\% \\
\bottomrule
\end{tabular}
\end{center}
\caption{Per-layer diagnostics at layer 15, step 950 of the main 1000-step seed-42 runs (per-layer breakdown in Appendix Figure~\ref{fig:diagnostics}). At the tested learning rates, Muon grows weights $2.2\times$ faster, producing ${\sim}8\times$ more saturation. EMA-blend has Muon-scale weights but keeps the post-blend erf argument in a low-saturation regime via $\hat{\sigma}$.}
\label{tab:weight_growth}
\end{table}

At the main factorial's learning rates, Muon produces weight norms of $\sim$156 vs.\ AdamW's $\sim$70 ($2.2\times$) by step 950. The faster weight growth drives erf input std from 1.8 (AdamW) to 11.8 (Muon), pushing saturation from 10\% to 83\%.\footnote{For vanilla Derf and AdamW, we define saturation as the fraction of elements where $|\mathrm{erf}(\alpha x + s)| > 0.99$. For EMA-blend, we report saturation of the actual post-blend erf argument.} EMA-blend retains Muon-scale weights ($\sim$164) but reduces the post-blend erf input std to 1.3, with only 2\% post-blend saturation. The key distinction is therefore not weight growth alone, but whether the normalizer still measures and corrects the activation scale seen by erf.

\subsection{$\alpha=0.3$ Supports the Saturation Mechanism}
\label{sec:alpha03}

Lowering Derf's $\alpha$ from its published default (0.5) to 0.3 widens erf's useful regime, keeping more elements in the near-linear region where $\mathrm{erf}(\alpha x) \approx 2\alpha x/\sqrt{\pi}$ approximately preserves relative magnitude. At $\alpha\!=\!0.3$, Derf+Muon reaches 3.511\,{\scriptsize$\pm$\,0.002} (3 seeds), recovering ${\sim}80\%$ of the 0.97-nat gap to RMSNorm+Muon (Appendix Table~\ref{tab:multiseed}). The interaction flips sign: $-$0.175 (95\% CI $[-0.19, -0.15]$) vs.\ $+$0.659 at $\alpha\!=\!0.5$. Under AdamW, $\alpha\!=\!0.3$ is slightly worse than 0.5 (4.248 vs.\ 4.192), so the benefit is specific to the Muon pairing. This $\alpha$ is not the published default of \citet{chen2026strongernormalizationfreetransformers}; identifying it as a Muon-compatible operating point in the tested regime is a finding of this work.

\subsection{Per-Layer and Temporal Patterns}

\paragraph{Across layers.} Early layers (L0--L3) are unsaturated under all configs; deep layers (L12--L15) diverge sharply between Muon and AdamW (Figure~\ref{fig:diagnostics}). EMA-blend weights grow at Muon rates ($\|W\|_F \approx 164$) yet post-blend erf input std stays at $\sim$1.3 (vs.\ 11.8 for vanilla Derf+Muon), because $\hat{\sigma}$ absorbs the activation growth.

\begin{figure}[t]
\centering
\includegraphics[width=0.76\columnwidth]{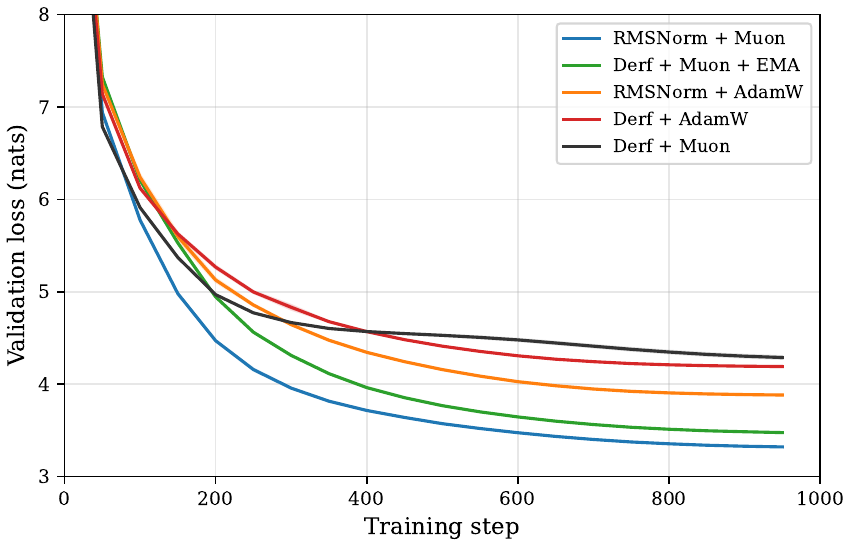}
\caption{Validation loss over 1000 steps. EMA-blend ($\lambda\!=\!0.9$) converges to within 0.15 nats of RMSNorm+Muon while outperforming RMSNorm+AdamW by 0.41 nats. Derf+Muon fails to outperform Derf+AdamW. Multi-seed variance is negligible. Three-seed results appear in Appendix~\ref{app:multiseed}.}
\label{fig:loss_curves}
\end{figure}

\paragraph{Over training.} Under Muon, saturation rises monotonically to ${\sim}80\%$. Under AdamW, it peaks early then recedes as $\sqrt{v_t}$ dampens growth (Figure~\ref{fig:sat_over_time}). EMA-blend post-blend saturation drops below 10\% once $\hat{\sigma}$ converges and reaches ${\sim}1\%$ by late training (Figures~\ref{fig:norm_out_var}--\ref{fig:sat_over_time} in Appendix~\ref{app:figures}). Notably, Derf+Muon trains to completion with no NaN or gradient explosion; the degradation is visible only as a widening loss gap from step ${\sim}200$ onward (Figure~\ref{fig:loss_curves}), meaning short pilot runs would miss it.

\subsection{Ruling Out Alternative Explanations}
\label{sec:ablations}

We test five alternative explanations for the Derf-Muon failure (Table~\ref{tab:fixes}; implementation details in Appendix~\ref{app:fixes}). Four interventions targeting gradients, weight norms, alpha adaptation, and growth rate fail to recover quality. Only desaturation (asinh) and scale restoration (EMA-blend) help, and their 35~pp recovery gap despite similar erf input distributions (std 1.08 vs.\ 1.19) is consistent with scale information being a distinct factor beyond desaturation alone.

\begin{table}[t]
\begin{center}
\small
\begin{tabular}{llll}
\toprule
\textbf{Hypothesis} & \textbf{Intervention} & \textbf{Outcome} & \textbf{Conclusion} \\
\midrule
Large gradients & Grad clipping & Silent degrad. (gap 0.97) & Not the cause \\
Weight norms alone & Weight decay & No effect (gap 0.97) & Too weak \\
Alpha can self-correct & Alpha LR $\uparrow$ & No effect (gap 0.97) & Ineffective in saturation regime \\
Growth rate alone & Residual scaling & Diverges & Insufficient recovery \\
Saturation alone & asinh compress. & 49\% recovery (gap 0.50) & Helps, not sufficient \\
Missing scale info. & EMA-blend & 84\% recovery (gap 0.15) & Saturation + scale \\
\bottomrule
\end{tabular}
\end{center}
\caption{Alternative-explanation tests (targeted probes). Rows 1--4 and asinh are single-seed probes; EMA-blend is replicated with 3 seeds. Only EMA-blend, which restores scale information, achieves substantial recovery. The 35~pp gap between asinh and EMA-blend (similar erf input distributions, different recovery) is consistent with scale information being a distinct factor. Percentages are relative to the full Derf+Muon gap of $+$0.97 nats.}
\label{tab:fixes}
\end{table}

\section{EMA-Blend as a Diagnostic Intervention}
\label{sec:ema}

EMA-blend simultaneously restores scale and reduces saturation (83\% to 2\% post-blend); the asinh comparison (Section~\ref{sec:ablations}) partially disentangles these effects. At 1000 steps, $\alpha\!=\!0.3$ achieves comparable recovery (${\sim}80\%$, Section~\ref{sec:alpha03}); EMA-blend's additional value lies in its adaptive scaling: the effective $\alpha_{\mathrm{eff}} = \alpha(1\!-\!\lambda + \lambda/\hat{\sigma})$ automatically shrinks as residual magnitudes grow, a property that fixed $\alpha$ lacks and that may prove necessary at longer training horizons (Section~\ref{sec:limitations}).

At $\lambda\!=\!0.9$, EMA-blend reaches \textbf{3.476} {\scriptsize $\pm$ 0.001} at 1000 steps, leaving a gap of just 0.15 nats to RMSNorm+Muon (vs.\ 0.97 for vanilla Derf+Muon) and recovering 84\% of the Derf penalty. Together with the $\alpha\!=\!0.3$ result, this is the strongest evidence for the proposed mechanism.

The method is normalization-amortized, not normalization-free: each site caches one scalar $\hat{\sigma}$, refreshed once per step via $\hat{\sigma}_t = (1\!-\!m)\hat{\sigma}_{t-1} + m \cdot \mathrm{std}(x_t)$. Since $\hat{\sigma}$ is scalar, the blend simplifies algebraically:
\begin{equation}
    (1\!-\!\lambda)\,x + \lambda\,\frac{x}{\hat{\sigma}} \;=\; x\!\left(1 - \lambda + \frac{\lambda}{\hat{\sigma}}\right)
    \label{eq:blend_simplify}
\end{equation}
so the forward pass reduces to $\mathrm{erf}(\alpha_{\mathrm{eff}} \cdot x + s)$ where $\alpha_{\mathrm{eff}} = \alpha(1 - \lambda + \lambda/\hat{\sigma})$, identical per-element cost to vanilla Derf. Table~\ref{tab:p_ablation} shows the effect of varying $\lambda$.

\begin{table}[t]
\begin{center}
\begin{tabular}{lcc}
\toprule
\textbf{$\lambda$} & \textbf{Val Loss} & \textbf{Gap to RMS+Muon} \\
\midrule
0.0 (vanilla Derf) & 4.289 {\scriptsize $\pm$ 0.017} & $+$0.967 \\
0.5 & 3.588 {\scriptsize $\pm$ 0.060} & $+$0.267 \\
0.7 & 3.504 {\scriptsize $\pm$ 0.003} & $+$0.182 \\
\textbf{0.9} & \textbf{3.476} {\scriptsize $\pm$ 0.001} & \textbf{$+$0.155} \\
1.0 & 6.64 {\scriptsize $\pm$ 0.001} (plateau) & $+$3.318 \\
\bottomrule
\end{tabular}
\end{center}
\caption{EMA-blend sweep over $\lambda$ at 1B, 1000 steps. Performance improves monotonically from $\lambda\!=\!0.5$ to $\lambda\!=\!0.9$, then collapses at $\lambda\!=\!1.0$. $\lambda\!=\!0.9$ is best; both $\lambda\!=\!0.7$ and $\lambda\!=\!0.9$ are replicated with 3 seeds.}
\label{tab:p_ablation}
\end{table}

All values in $\lambda \in [0.5, 0.9]$ substantially reduce the gap, with $\lambda\!=\!0.9$ best (gap 0.155, 3-seed avg). At $\lambda\!=\!1.0$, performance collapses (Appendix~\ref{app:p1}): the $(1\!-\!\lambda)$ raw component acts as a necessary growth brake, and in our $\lambda$ sweep, 10\% raw signal was the smallest tested fraction that remained effective.

\begin{table}[t]
\begin{center}
\begin{tabular}{lccc}
\toprule
\textbf{Method} & \textbf{Allreduces / step} & \textbf{Relative} & \textbf{Quality} \\
\midrule
RMSNorm & 33 $\times$ 64 $\times$ 2 = 4,224 & $1\times$ & best \\
Derf + EMA & 33 $\times$ 1 = 33 & $\sim$0.008$\times$ & $-$0.15 nats \\
DyT & 0 & $0\times$ & $-$0.22 nats \\
Derf (no EMA) & 0 & $0\times$ & $-$0.97 nats \\
\bottomrule
\end{tabular}
\end{center}
\caption{Theoretical per-step allreduce \emph{operation} count under tensor parallelism (forward $+$ backward). The $128\times$ reduction in operation count (4{,}224 vs.\ 33) translates to $7.8\times$ measured norm-layer wall-clock speedup at 8-way TP (Figure~\ref{fig:tp_benchmark}); the gap between theoretical and measured reflects allreduce fusion and overlap in modern TP implementations. DyT and vanilla Derf are fully reduction-free, but only DyT preserves quality with Muon.}
\label{tab:reductions}
\end{table}

\begin{figure}[t]
\centering
\includegraphics[width=0.76\columnwidth]{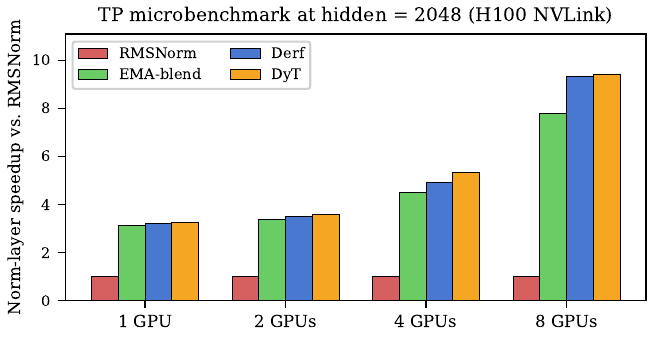}
\caption{Tensor parallelism benchmark on H100 NVLink (1/2/4/8 GPUs; training uses a single H200, Section~\ref{sec:setup}). Norm-layer wall-clock time at hidden$=$2048. DyT and Derf are fully reduction-free; EMA requires 33 allreduces/step vs.\ RMSNorm's 4,224. At 8-way TP: DyT $9.4\times$, Derf $9.3\times$, EMA $7.8\times$ faster than RMSNorm. At 1-GPU (no communication), the gap is pure compute: ${\sim}3.2\times$ for all three, confirming the speedup scales with TP degree.}
\label{fig:tp_benchmark}
\end{figure}

\section{Discussion}
\label{sec:discussion}

We do not claim that Muon is generally incompatible with bounded normalizers; the DyT control argues against that. Our claim is narrower: Derf's published operating point interacts poorly with Muon's scale dynamics in this regime, and the failure produces no crash signal. DyT and Derf behave similarly under AdamW, yet they separate sharply under Muon because Muon magnifies Derf's specific saturation geometry.

\paragraph{Fragility vs.\ interaction.}
One could frame our finding as a Derf robustness failure rather than an interaction. We retain the interaction framing because the failure is pairing-specific (DyT+Muon works despite comparable weight growth; Table~\ref{tab:dyt_diagnostics}), AdamW's $\sqrt{v_t}$ partially self-corrects where Muon cannot (Figure~\ref{fig:sat_over_time}), and the practical takeaway (normalizer choice is not independent of optimizer choice) holds regardless.

\paragraph{Muon LR ablation.}
A single-seed LR probe is directionally consistent: lowering Muon LR from 0.02 to 0.01 reduces the Derf gap from $+$0.95 to $+$0.28; the non-monotonic result at 0.005 ($+$0.38) reflects undertraining (Appendix Table~\ref{tab:lr_sweep}).

\paragraph{$\alpha$ sensitivity.}
The $\alpha\!=\!0.3$ result (Section~\ref{sec:alpha03}) strongly supports the proposed mechanism; details and numbers appear there.

\paragraph{Practical implication.} RMSNorm+Muon remains the strongest configuration at 1B. For practitioners pursuing reduction-free training, DyT+Muon preserves most of Muon's advantage with zero allreduces. If Derf is preferred, lowering $\alpha$ to 0.3 recovers ${\sim}80\%$ of the gap at zero overhead, while EMA-blend recovers ${\sim}84\%$ with adaptive scaling that may prove necessary at longer horizons (Section~\ref{sec:limitations}). In this regime, normalization-free methods benefit from optimizer-aware validation rather than transfer from AdamW results alone.

\section{Related Work}
\label{sec:related}

\paragraph{Normalization-free transformers.}
T-Fixup \citep{pmlr-v119-huang20f} trains transformers without normalization via careful initialization and scaling. DyT \citep{Zhu_2025_CVPR} uses $\tanh(\alpha x)$, matching RMSNorm at 34B+ but showing a gap at smaller scales. Derf \citep{chen2026strongernormalizationfreetransformers} uses $\mathrm{erf}(\alpha x + s)$. Both were tested exclusively with AdamW; we study both with Muon and find the coupling is Derf-specific (Section~\ref{sec:results}).

\paragraph{The Muon optimizer.}
Muon \citep{jordan2024muon} orthogonalizes the gradient via Newton-Schulz iterations \citep{bernstein2025deriving}. Moonlight \citep{liu2025muonscalablellmtraining} scaled Muon to 3B/16B MoE; SOAP \citep{vyas2025soap} and MuonClip \citep{kimiteam2026kimik2openagentic} address stability. All assume standard normalization layers.

\paragraph{Normalization-optimizer interactions.}
Normalization placement \citep{pmlr-v119-xiong20b}, depth-dependent scaling \citep{wang2022deepnetscalingtransformers1000}, and Adam's $\sqrt{v_t}$ \citep{DBLP:journals/corr/KingmaB14, ba2016layernormalization} all affect training stability. Most closely related, \citet{NEURIPS2024_986292a9} show that optimizer and normalization jointly affect outlier features, and \citet{NEURIPS2024_c04d37be} show normalization can mask effective-learning-rate dynamics. We focus specifically on Muon paired with bounded normalization-free layers, showing the interaction is selective to Derf and using EMA-blend as a scale-restoring probe.

\section{Limitations}
\label{sec:limitations}

\begin{itemize}
    \item \textbf{Scale and tokens.} We test at 1B parameters, ${\sim}$1B tokens (vs.\ ${\sim}$20B Chinchilla-optimal \citep{NEURIPS2022_c1e2faff}). Whether the interaction persists at 7B+ and longer schedules remains open.
    \item \textbf{EMA hyperparameters.} We sweep $\lambda \in \{0.5, 0.7, 0.9, 1.0\}$ but fix $m = 0.5$; sweeping momentum may further improve results.
    \item \textbf{Fixed $\alpha$ may not scale.} The $\alpha\!=\!0.3$ fix works at 1000 steps because residual magnitudes are still moderate. At Chinchilla-optimal ${\sim}$20K steps, continued weight growth under Muon (${\sim}2.2\times$ faster) may push residuals beyond the current regime, and a fixed $\alpha$ may prove insufficient to simultaneously provide useful nonlinearity early and resist saturation late. This motivates adaptive mechanisms such as EMA-blend for longer horizons.
    \item \textbf{Scope.} TP benchmark is intra-node only; LlamaForCausalLM and standard Muon only. Other architectures, Muon variants, and cross-node TP are untested.
\end{itemize}

\section{Conclusion}
\label{sec:conclusion}

At 1B parameters and 1000 steps, Derf shows a large silent interaction with Muon at published defaults ($+$0.66) while DyT does not ($-$0.10). Reducing $\alpha$ from 0.5 to 0.3 recovers ${\sim}80\%$ of the gap, providing strong evidence that erf's narrow operating regime under Muon is a primary driver of the failure; EMA-blend recovers ${\sim}84\%$ via adaptive scaling that may prove necessary at longer horizons. In this regime, normalization-free methods benefit from joint validation with the target optimizer; behavior under AdamW does not guarantee behavior under spectral updates.


\section*{Reproducibility Statement}

All code, training scripts, and hyperparameters are in the supplementary material. We use FineWeb-Edu and LlamaForCausalLM (HuggingFace Transformers). Training runs use a single NVIDIA H200 SXM; the TP microbenchmark uses H100 NVLink (1/2/4/8 GPUs). W\&B run IDs are provided.


\bibliography{colm2026_conference}
\bibliographystyle{colm2026_conference}

\appendix

\section{Multi-Seed Results}
\label{app:multiseed}

\begin{table}[H]
\begin{center}
\begin{tabular}{lcccc}
\toprule
\textbf{Config} & \textbf{s42} & \textbf{s43} & \textbf{s44} & \textbf{Avg} \\
\midrule
RMSNorm+Muon & 3.321 & 3.319 & 3.325 & \textbf{3.322} \\
RMSNorm+AdamW & 3.874 & 3.888 & 3.888 & \textbf{3.883} \\
DyT+Muon & 3.545 & 3.538 & 3.541 & \textbf{3.541} \\
DyT+AdamW & 4.204 & 4.192 & 4.207 & \textbf{4.201} \\
EMA-blend ($\lambda$=0.9) & 3.476 & 3.476 & 3.477 & \textbf{3.476} \\
EMA-blend ($\lambda$=0.7) & 3.502 & 3.503 & 3.507 & \textbf{3.504} \\
Derf+AdamW & 4.189 & 4.192 & 4.194 & \textbf{4.192} \\
Derf+Muon & 4.271 & 4.306 & 4.290 & \textbf{4.289} \\
\midrule
Derf ($\alpha$=0.3)+Muon & 3.513 & 3.510 & 3.510 & \textbf{3.511} \\
Derf ($\alpha$=0.3)+AdamW & 4.259 & 4.228 & 4.256 & \textbf{4.248} \\
\bottomrule
\end{tabular}
\end{center}
\caption{Validation loss at step 1000 across seeds. All configs have 3 seeds. Std $\leq$ 0.017.}
\label{tab:multiseed}
\end{table}

\section{Statistical Summary}
\label{app:stats}

Contrasts computed as mean $\pm$ sample std over 3 seeds per cell. Bootstrap CIs (10,000 resamples) are also shown but should be interpreted as lower bounds on uncertainty given $n\!=\!3$.

\begin{table}[H]
\begin{center}
\begin{tabular}{lcc}
\toprule
\textbf{Contrast} & \textbf{Mean} & \textbf{95\% CI} \\
\midrule
Derf gap (Muon) & $+$0.967 & $[0.95, 0.98]$ \\
Derf gap (AdamW) & $+$0.308 & $[0.30, 0.32]$ \\
Derf interaction & $+$0.659 & $[0.64, 0.68]$ \\
Ratio (Muon/AdamW gap) & $3.1\times$ & $[3.0, 3.2]$ \\
\midrule
DyT gap (Muon) & $+$0.220 & $[0.21, 0.23]$ \\
DyT gap (AdamW) & $+$0.318 & $[0.30, 0.33]$ \\
DyT interaction & $-$0.098 & $[-0.12, -0.08]$ \\
\midrule
EMA ($\lambda\!=\!0.9$) gap & $+$0.155 & $[0.15, 0.16]$ \\
EMA ($\lambda\!=\!0.7$) gap & $+$0.182 & $[0.18, 0.19]$ \\
Derf+Muon vs Derf+AdamW & $+$0.097 & $[0.08, 0.11]$ \\
\midrule
Derf $\alpha\!=\!0.3$ gap (Muon) & $+$0.189 & $[0.18, 0.19]$ \\
Derf $\alpha\!=\!0.3$ gap (AdamW) & $+$0.364 & $[0.34, 0.39]$ \\
Derf $\alpha\!=\!0.3$ interaction & $-$0.175 & $[-0.19, -0.15]$ \\
\bottomrule
\end{tabular}
\end{center}
\caption{Key contrasts with bootstrap 95\% CIs (10K resamples, $n\!=\!3$/cell). CIs are lower bounds on true uncertainty given small $n$.}
\label{tab:stats}
\end{table}

\section{Failed Interventions: Details}
\label{app:fixes}

\paragraph{Gradient clipping.} Bounds gradient norms but loss gap continues widening. Problem is information loss, not gradient magnitude.

\paragraph{Weight decay (0.01).} Shrinks weights 0.02\%/step, too slow against 16 layers of compounding growth.

\paragraph{Alpha LR $\uparrow$ (0.1).} Alpha gradient $\propto \exp(-(\alpha x)^2)$ is mathematically zero in the saturation zone regardless of step size.

\paragraph{Residual scaling.} Per-sublayer scalars (init 0.5) learn differentiation (early layers $\to$ 0.43, late attention $\to$ 0.53) but cannot self-regulate fast enough; in the targeted probe, loss reverses by step 400 (Table~\ref{tab:resscale}).

\begin{table}[H]
\begin{center}
\begin{tabular}{lccc}
\toprule
\textbf{Layer} & \textbf{attn} & \textbf{ffn} & \textbf{Interpretation} \\
\midrule
0 & 0.50$\to$0.43 & 0.50$\to$0.42 & Early shrinks \\
7 & 0.50$\to$0.50 & 0.50$\to$0.47 & Middle stable \\
15 & 0.50$\to$0.53 & 0.50$\to$0.45 & FFN shrinks \\
\bottomrule
\end{tabular}
\end{center}
\caption{Residual scaling parameters at step 400 in the targeted probe.}
\label{tab:resscale}
\end{table}

\paragraph{Asinh pre-compression.} $\mathrm{erf}(\alpha \cdot \mathrm{asinh}(\beta x)/\beta + s)$ with learnable $\beta$. Post-asinh saturation stays at 0\% at all layers; the actual erf argument has std 1.46 at L15 step 340 and 1.08 at step 900, comparable to EMA-blend's 1.66 and 1.19 at the same steps. Alpha gradients are $2\times$ stronger than vanilla Derf+Muon. The 400-step warmup-200 diagnostic run ends at 4.42, close to vanilla Derf at the same horizon, but a longer single-seed follow-up run reaches step 936 with last validation logged at step 900, where it improves materially over vanilla Derf+Muon (3.82 vs.\ 4.29). Even so, it remains well above RMSNorm+Muon (3.33 at step 900) and EMA-blend (3.51 at step 900), so desaturating erf helps but does not close the gap.

\section{EMA-Blend $\lambda = 1.0$}
\label{app:p1}

Full normalization ($\lambda = 1.0$) caused activation RMS at L15 to reach 7,214 during training. Loss plateaued at 6.64. The $(1-\lambda)$ raw component (10\% at $\lambda\!=\!0.9$) acts as a necessary growth brake.

\section{DyT vs.\ Derf Diagnostics}
\label{app:dyt_diagnostics}

\begin{table}[H]
\begin{center}
\begin{tabular}{lcccc}
\toprule
\textbf{Metric (L15, step 999)} & \textbf{DyT+Muon} & \textbf{DyT+AdamW} & \textbf{Derf+Muon} & \textbf{Derf+AdamW} \\
\midrule
MLP weight $\|W\|_F$ & 144 & 70 & 156 & 70 \\
Nonlinearity input std & 2.96 & 1.92 & 10.65 & 1.76 \\
Saturated (\%) & 27\% & 12\% & 79\% & 9\% \\
FFN $\alpha$ (init $\to$ final) & 0.30$\to$0.175 & 0.30$\to$0.40 & 0.50$\to$0.36 & 0.50$\to$0.44 \\
\bottomrule
\end{tabular}
\end{center}
\caption{DyT vs.\ Derf diagnostics at layer 15, step 999 (seed 42). Despite comparable Muon-driven weight growth, DyT has far lower saturation because tanh has a wider useful regime and DyT's $\alpha$ adapts more aggressively.}
\label{tab:dyt_diagnostics}
\end{table}

\section{125M Results}
\label{app:125m}

\begin{table}[H]
\begin{center}
\begin{tabular}{lccc}
\toprule
\textbf{Run} & \textbf{Steps} & \textbf{Val Loss} & \textbf{Muon grad norm} \\
\midrule
RMSNorm+Muon & 999 & 3.79 & 0.066 \\
RMSNorm+AdamW & 999 & 3.89 & n/a \\
DyT+Muon ($\alpha=0.5/0.3$) & 999 & 4.38 & 0.089 \\
Derf+Muon ($\alpha=0.5$) & 186$^*$ & 6.07 & 0.184 \\
\bottomrule
\end{tabular}
\end{center}
\caption{Exploratory 125M pilot results. DyT+Muon shows a moderate gap ($+$0.59) at 125M; at 1B, DyT+Muon gap narrows to $+$0.22 (Table~\ref{tab:factorial}), consistent with DyT scaling better than Derf under Muon. $^*$Derf+Muon terminated by dataset timeout, not divergence, so the 125M pilot is not used to draw scale-dependent conclusions about the Derf interaction.}
\label{tab:125m-full}
\end{table}

\section{Muon LR Probe}
\label{app:lr_probe}

\begin{table}[H]
\begin{center}
\begin{tabular}{lccc}
\toprule
\textbf{Muon LR} & \textbf{RMSNorm} & \textbf{Derf} & \textbf{Derf gap} \\
\midrule
0.02 & 3.321 & 4.271 & $+$0.950 \\
0.01 & 3.379 & 3.656 & $+$0.277 \\
0.005 & 3.577 & 3.954 & $+$0.377 \\
\bottomrule
\end{tabular}
\end{center}
\caption{Muon LR probe (val loss at step 1000; single seed s42). Lowering LR from 0.02 to 0.01 reduces the Derf gap, consistent with slower weight growth producing less saturation. The non-monotonic 0.005 result reflects undertraining.}
\label{tab:lr_sweep}
\end{table}

\section{Additional Figures}
\label{app:figures}

\begin{figure}[H]
\centering
\includegraphics[width=\columnwidth]{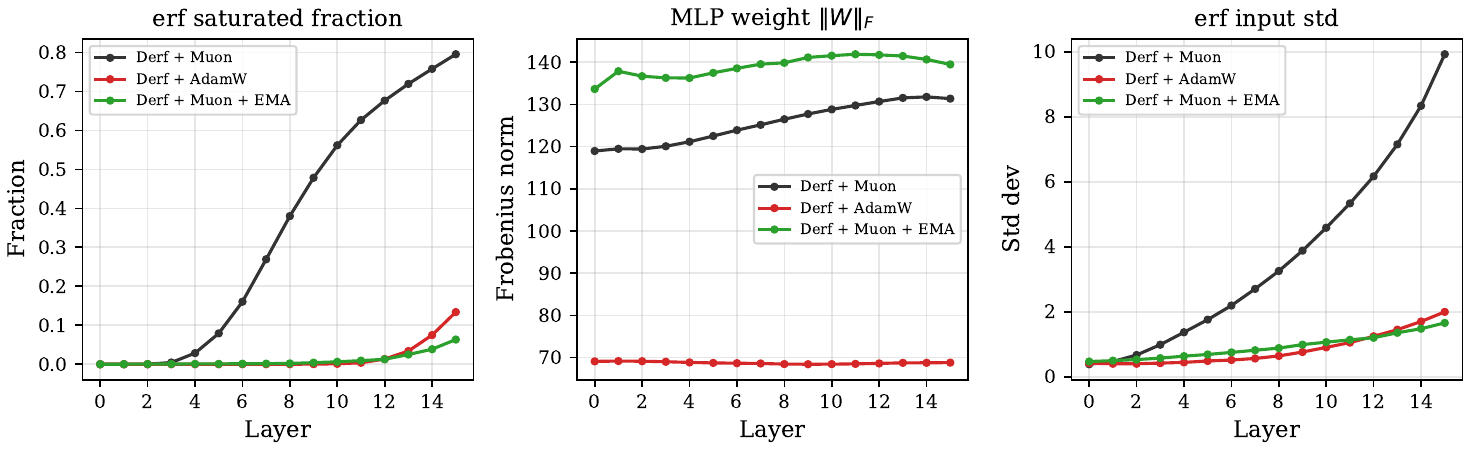}
\caption{Per-layer diagnostics at step 340 of the main 1000-step seed-42 runs. Saturation, weight growth, and erf input magnitude increase with depth. For EMA-blend, the saturation panel uses post-blend saturation of the actual erf argument. Muon drives the vanilla Derf metrics higher than AdamW, while EMA-blend decouples weight growth from erf input.}
\label{fig:diagnostics}
\end{figure}

\begin{figure}[H]
\centering
\includegraphics[width=\columnwidth]{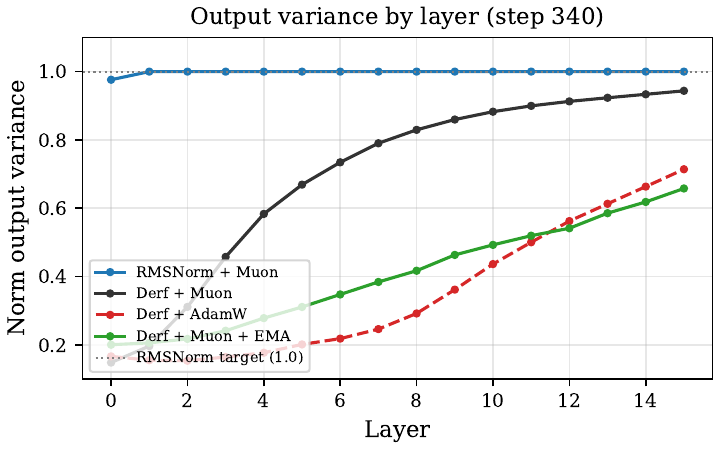}
\caption{Norm output variance by layer at step 340 of the main 1000-step seed-42 runs. Derf+AdamW and EMA-blend lines begin at layer ${\sim}7$ because earlier layers have variance $< 0.4$, below the plotted range. Counterintuitively, Derf+Muon has the \emph{highest} variance among Derf configs (0.94 at L15). This is not because it preserves information, but because 80\% saturation pushes outputs to $\pm 1$, a bimodal distribution with high variance but substantially reduced representational resolution.}
\label{fig:norm_out_var}
\end{figure}

\begin{figure}[H]
\centering
\includegraphics[width=\columnwidth]{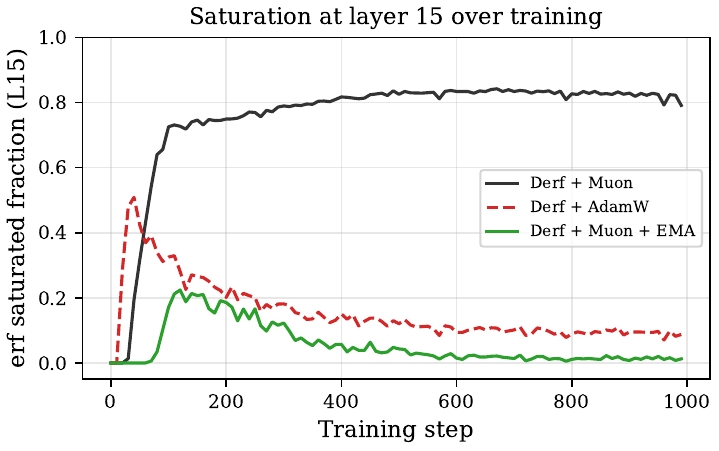}
\caption{erf saturation at layer 15 over 1000 training steps. Derf+Muon saturates rapidly and stays at ${\sim}80\%$. Derf+AdamW peaks at ${\sim}50\%$ around step 50 then recedes as AdamW's adaptive $\sqrt{v_t}$ dampens growth, evidence that AdamW also fights Derf's saturation, but wins the fight. For EMA-blend (labeled ``post-blend''), we plot saturation of the actual erf argument \emph{after} blending; it falls below 10\% once $\hat{\sigma}$ converges and reaches ${\sim}1\%$ by late training.}
\label{fig:sat_over_time}
\end{figure}

\begin{figure}[H]
\centering
\includegraphics[width=\columnwidth]{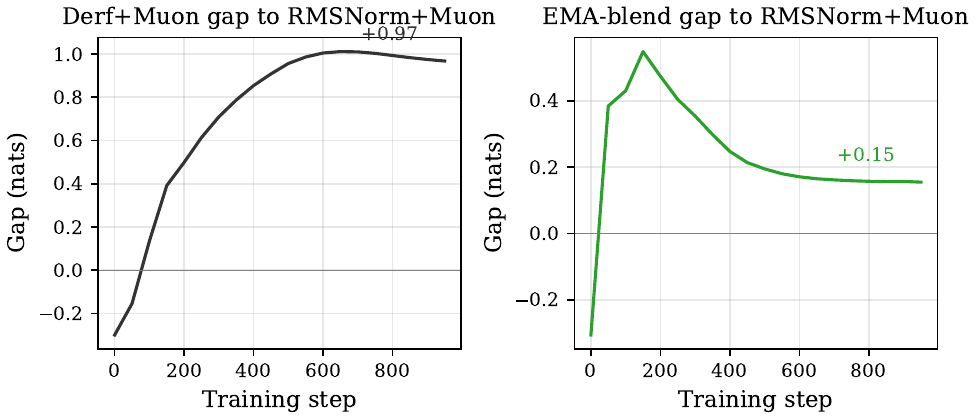}
\caption{Gap trajectories (1000-step cosine schedule). Left: Derf+Muon's gap widens monotonically and plateaus. Right: EMA-blend's gap peaks at $+$0.53 around step ${\sim}175$ (before $\hat{\sigma}$ converges) then narrows. Both trends are partially confounded with cosine decay.}
\label{fig:gap_trajectories}
\end{figure}

\end{document}